\documentclass[10pt,twocolumn,letterpaper,final,pagenumbers]{article}
\usepackage[pagenumbers]{cvpr} 
\usepackage{graphicx}
\usepackage{amsmath}
\usepackage{amssymb}
\usepackage{booktabs}
\usepackage{algorithm}
\usepackage{algpseudocode}
\usepackage{amsmath} 
\usepackage{multirow}
\usepackage[T5]{fontenc}   
\usepackage[utf8]{inputenc} 
\usepackage{lipsum}
\usepackage{array}

\usepackage[pagebackref,breaklinks,colorlinks]{hyperref}

\usepackage[capitalize]{cleveref}
\crefname{section}{Sec.}{Secs.}
\Crefname{section}{Section}{Sections}
\Crefname{table}{Table}{Tables}
\crefname{table}{Tab.}{Tabs.}

\usepackage{fancyhdr}
\fancypagestyle{page2style}{
    \fancyhf{}
    \fancyfoot[L]{Special comment for Page 2} 
}

\fancypagestyle{page3style}{
    \fancyhf{}
    \fancyfoot[L]{* Gx is a x-axis of coordinate \\ 
                  * Gy is a y-axis of coordinate} 
}

\fancypagestyle{page5style}{
    \fancyhf{}
    \fancyfoot[L]{* model1 is model detection \\ 
                  * model2 is model classify} 
}

\begin{document}

\title{Detection Fire in Camera RGB-NIR}

\author{
Nguyễn Trường Khải\quad \quad Lương Đức Vinh  \\ 
{\tt\small khainguyentruong@gmail.com \quad \quad vinhld@viettel.com}
}

\maketitle

\begin{abstract}
    \noindent Improving the accuracy of fire detection for infrared night vision cameras has long been a challenging problem. Previous studies have achieved high accuracy using popular detection models, such as YOLOv7 with an input image size of $(640 \times 1280)$ achieving an accuracy of $\text{mAP}_{50:95} = 0.51$, RT-DETR with $\text{mAP}_{50:95} = 0.65$ (image size: $640 \times 640$), and YOLOv9 with $\text{mAP}_{50:95} = 0.598$ (image size: $640 \times 640$). However, due to limited resources for dataset creation, some issues persist, particularly the misclassification of bright lights as fire. 
    This report contributes three main parts: an additional NIR dataset, a two-stage model, and Patched-YOLO.\textbf{First, the NIR Dataset:} Due to the limited data in the previous dataset, we have researched and experimented with various methods to augment two main datasets: the NIR dataset and the classification dataset. \textbf{Second, the Two-Stage Model:} To enhance the accuracy of fire detection using infrared cameras at night while minimizing false positives caused by artificial lights, we propose a two-stage approach utilizing two key models: YOLOv11 and EfficientNetV2-B0. With the proposed pipeline, we are confident that the detection accuracy surpasses previous fire detection models. Specifically, for the fire detection task at night.\textbf{Third, Patched-YOLO:} Fire detection in RGB images poses challenges, particularly in detecting small objects and those located far from the camera. To address this issue, we propose using \textit{Patched-YOLO} to enhance the model's detection capabilities. Details of our contributions will be discussed in the following sections.  
\end{abstract}

\section{Introduction}
\label{sec:intro}
\subsection{Background \& Motivation}  
\noindent Fire detection is a crucial task in ensuring public safety, especially for nighttime fire incidents, where visibility is significantly reduced. According to the Vietnam Fire and Rescue Police Department, the number of fire-related incidents has risen sharply in recent years. In 2024 alone, Vietnam recorded 2,222 fire cases, leading to severe casualties and financial damages estimated at 127.9 billion VND \cite{pccc}.  

Traditional fire detection systems rely on visible spectrum cameras, which face severe limitations in nighttime conditions due to poor lighting. Infrared (NIR) cameras provide an alternative solution by capturing thermal signatures of fire, enabling better detection in low-light environments. However, existing fire detection models still struggle with false positives, as bright artificial lights (e.g., streetlights, neon signs, and reflections) often resemble fire in infrared imagery.  
\subsection{Challenges in Nighttime Fire Detection and RGB images}  
\noindent Despite advancements in deep learning and computer vision, nighttime fire detection using NIR cameras presents several challenges:  

\begin{itemize}  
    \item \textbf{Lack of High-Quality Datasets}: Existing fire detection datasets mostly focus on visible light images, while publicly available NIR fire datasets remain scarce. 
    
    \item \textbf{False Positives}: Bright lights from street lamps, vehicles, and other sources often get misclassified as fire.  
 
    \item \textbf{Detecting small distant fire in images}: Small objects that are far from the camera are always a challenge for detection models, and this model is no exception.  
\end{itemize}  

\subsection{Our Contributions} 
\noindent To address the challenges of nighttime fire detection, we propose a some solution:
\begin{itemize} 
    \item \textbf{Nighttime Fire Dataset (NIR + Classification Dataset):}A newly designed dataset specifically for infrared-based fire detection at night, rigorously tested and improved. 
    
    \item \textbf{Enhanced Edge Detection (Sobel and Laplacian):}Extracting distinguishing features between fire and other bright objects to reduce false positives.    

    \item \textbf{YOLOv11n for Fire Detection:} A state-of-the-art object detection model optimized for infrared-based fire detection, demonstrating superior performance. 
    
    \item \textbf{EfficientNetV2-B0 for Fire Classification} – A lightweight yet powerful classification model to filter out false positives caused by artificial lights.
    
    \item \textbf{Patched-YOLO:} An improved detection approach that enhances the model’s ability to detect small and distant fires more effectively. 
 
\end{itemize}  

Our preliminary results retraining the YOLOv11n model on the dataset we added achieved an accuracy of $\text{AP}_{50:95} = 0.622$. The full evaluation of the two-stage pipeline will be conducted once the classification module is fully integrated.  


\section{Related Work}
\label{sec:formatting}
\noindent Fire detection using computer vision has been extensively explored in recent years. Traditional methods rely on color space analysis and Gaussian Mixture models or image processing techniques to detect fire \cite{gaussian, 6}. However, these approaches only achieve high performance when images have high resolution and good lighting conditions.

Deep learning-based models, such as CNNs and popular object detection networks, have significantly improved fire detection accuracy. For example, experiments with YOLO models \cite{yolov7, yolov9, yolov11} have shown promising results in real-time fire detection, achieving high precision. Nevertheless, these models still misclassify strong light sources, such as streetlights or indoor lighting at night, as fire, especially in infrared imagery.

Methods that rely solely on object detection models or individual image processing techniques have their own limitations. To address these issues, we propose a two-stage approach that integrates object detection with edge-based feature extraction methods using Sobel and Laplacian filters. Our approach enhances accuracy in cases where the object detection model is uncertain, particularly in nighttime environments and distant fire spots where lighting conditions cause glare. However, this is only based on theoretical assumptions. We are currently in the process of validating it and will strive to provide a formal proof as soon as possible.

Datasets for nighttime fire detection using infrared (NIR) cameras are extremely scarce. During our research, we identified a dataset created by a group of French researchers~\cite{}. However, due to licensing restrictions, we were unable to use it directly. Instead, we utilized a previously published dataset from the same authors and applied various image processing techniques to convert RGB images into NIR representations, thereby improving the dataset. Additionally, we created a new classification dataset to enhance the model’s performance.  
Regarding object detection for distant fire instances, common methods involve dividing images into smaller patches, detecting objects within each patch, and then merging the results. However, these approaches face several challenges:  

\begin{itemize}
    \item If the object is located at the boundary of a patch, it may be split into multiple parts, preventing proper recognition.
    \item Small objects at long distances may lack sufficient pixel representation, leading to misdetections or omissions.
\end{itemize}
To mitigate these issues, we propose a novel \text{patched-YOLO} \cite{patched-yolo} approach, one method as the same the SaHi method\cite{SaHi} . 
This approach enhances detection reliability for small and distant fire objects, significantly improving model performance for nighttime infrared fire detection.



\section{Dataset}
\noindent Our proposed method utilizes two models to achieve the desired accuracy. Therefore, we employed two primary datasets: the \textbf{Dataset Detection NIR} and \textbf{Datasets Classify with Sobel}. 

\subsection{Dataset Detection NIR}
\noindent During the time we conducted this project, we received a dataset from our predecessors. However, the dataset was relatively small, with approximately 3,306 images for the training set and 286 images for the test set. Due to this data limitation, the model's accuracy reached its threshold. Therefore, we attempted to augment the training dataset using RGB-to-Grayscale conversion and Fusion GAN to generate additional NIR images from RGB images.
\subsection{Datasets Classify Sobel}
\noindent For the classification dataset, due to its specific nature, there is no publicly available dataset. As a result, we spent a significant amount of time experimenting with various image processing techniques to determine the best approach for dataset creation. The specific methods used to generate the dataset will be discussed in the \textbf{Method} section.



In summary, after applying various image processing techniques and enhancement methods, we successfully constructed two datasets tailored for the two models used in this study. Details about the number of images and bounding boxes for the detected objects can be found in Table \ref{table2}.

\begin{table}[h]
    \centering
    \renewcommand{\arraystretch}{1.5}
    \begin{tabular}{|c|c|c|c|}
        \hline
        
        {} & \textbf{Dataset} 
        & \textbf{Images} & \textbf{Bbox}  \\ \hline
        \multirow{2}{*}{\textbf{Dataset detect}} 
        & Train  & 26403 & 55261 \\ \cline{2-4}
        & Test   & 286 & 229 \\ \hline
        
        \multirow{4}{*}{\textbf{Dataset classify}} 
        & \multicolumn{3}{c|}{\textbf{Images}} \\ \cline{2-4}
        & \textbf{Dataset} & \textbf{Fire} & \textbf{No Fire} \\ \cline{2-4}
        & Train  & 6067 & 970 \\ \cline{2-4}
        & Test   & 2601 & 416 \\ \hline
    \end{tabular}
    \caption{Dataset statistic number of images and bounding boxes}
    \label{table2}
\end{table}
\section{ Method }
\subsection{Enhancing the Dataset}

\subsubsection{Dataset Detection NIR}
\noindent In the initial phase of our experiment, we used a dataset consisting of 3,306 images for training and 286 images for testing. These images were annotated by previous authors and originated from videos collected from YouTube, as well as real-world cases captured by surveillance cameras at gas stations, residential areas, and regions where fires had occurred or fire drills had been conducted. 

Due to the nature of the problem, which involves infrared camera images, the available datasets were extremely limited. To enrich the dataset and increase its diversity, we leveraged the RGB fire and smoke recognition dataset from FASDD \cite{FASDD}. We then filtered out images containing fire and converted the RGB images into NIR images to align with the problem requirements. To achieve this conversion, we experimented with various frameworks before deciding to focus on two primary ones.

\begin{figure}[h]
    \centering
    \includegraphics[width=\linewidth]{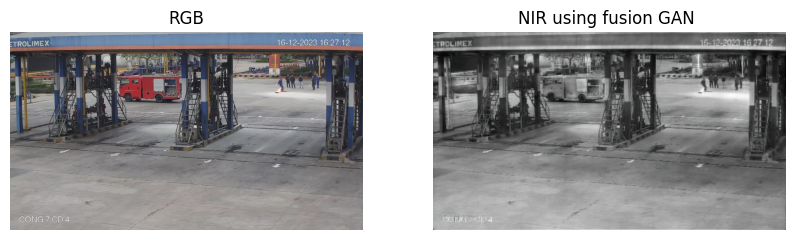}
    \caption{Conversion of RGB images to NIR using FIRE-GAN.}
    \label{fig2}
\end{figure}

Since there is no clear mathematical definition of NIR images, it is challenging to transform RGB images into NIR using a predefined formula. Infrared images are typically captured by cameras at night. However, with the advancement of generative AI, it is now possible to synthesize such images. \textbf{Our first} approach was to leverage generative AI to reconstruct the dataset. Several studies have explored methods to generate NIR images from RGB images using generative models. We adopted the FIRE-GAN model, proposed by J. F. Cipri\'an-S\'anchez et al. \cite{Ciprian2021}, which was trained on extensive wildfire datasets from France. The results of converting RGB images to NIR using FIRE-GAN are illustrated in Figure \ref{fig2}.

After filtering and converting the RGB images containing fire from the FASDD dataset \cite{FASDD} using FIRE-GAN, the entire process took approximately 48 hours. This significantly increased the dataset size, resulting in 28,721 images for training and 11,179 images for testing.

On the other hand, instead of using a model to convert RGB images to NIR, we proposed a \textbf{second} approach using a formula provided by the OpenCV library to convert images from RGB to grayscale. Through observations of NIR images captured by nighttime cameras and grayscale images, we found an 80-90\% similarity, as seen in Figure \ref{fig3}

\begin{figure}[h]
    \centering
    \includegraphics[width=\linewidth]{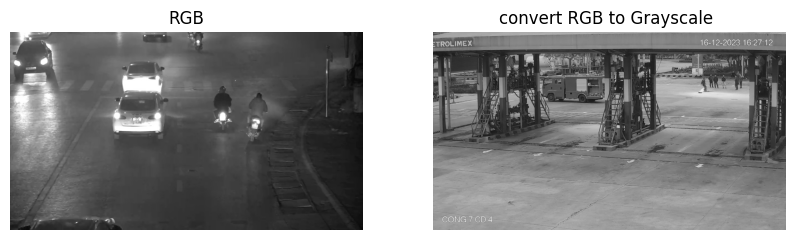}
    \caption{NIR same as image grayscale.}
    \label{fig3}
\end{figure}

Therefore, We use grayscale images because the mAP of the model trained with grayscale images is higher than the one trained with FIRE-GAN. After converting the filtered RGB images from the FASDD dataset \cite{FASDD} to grayscale, we added 25,415 images to the training set and 22,660 images to the test set compared to the previous dataset.



\subsubsection{Datasets Classify Sobel}
\thispagestyle{page3style}
\noindent For classification datasets, we focus on edge features of the object. Since infrared images can have varying color properties, shape and edge features remain consistent. Given that infrared camera images are often blurry, we applied the following image processing techniques to enhance dataset clarity.

Typically, to enhance object edges, primary methods such as Sobel and Laplacian operators are used. These methods apply convolution matrices to transform images into their edge representations. The specific formulas for the Sobel and Laplacian operators are given below:

\begin{equation}
    G_x =
    \begin{bmatrix} 
        -1 & 0 & 1 \\ 
        -2 & 0 & 2 \\ 
        -1 & 0 & 1 
    \end{bmatrix} * I, \quad 
    G_y =
    \begin{bmatrix} 
        -1 & -2 & -1 \\ 
         0 &  0 &  0 \\ 
         1 &  2 &  1 
    \end{bmatrix} * I \quad 
\end{equation}

\begin{equation}
    G = \sqrt{G_x^2 + G_y^2}
\end{equation}


Based on equations (1), it can be observed that the Sobel transformation operates by computing the first-order derivative in two directions: horizontal (X-axis) and vertical (Y-axis).  

First, we apply the Sobel filter along both the \( x \) and \( y \) axes. Then, we use the absolute value operation to normalize the gradient values. Finally, we apply the bitwise OR of cv2 lib operation to combine the results from both directions, producing a comprehensive edge-detected image.  
In practical applications, the Laplacian transform is performed through convolution between the image and a Laplacian kernel matrix, as shown in Equation (3).
\begin{equation}
    \nabla^2 I = \begin{bmatrix} 0 & 1 & 0 \\ 1 & -4 & 1 \\ 0 & 1 & 0 \end{bmatrix} * I
\end{equation}
Since applying edge detection filters such as Sobel and Laplacian can result in the loss of fine details and edge information, we incorporate a Gaussian filter to restore lost details and enhance the overall image quality. Following this step, we apply the Canny edge detector to precisely identify object boundaries. Finally, to effectively merge the extracted edge features, we employ a weighted combination of the processed images using the following formula:  

\begin{equation}
    I_{\text{final}} = I_{\text{laplacian}} + I_{\text{canny}} 
\end{equation}  

where \( I_{\text{laplacian}} \) represents the Laplacian-filtered image, \( I_{\text{canny}} \) is the Canny edge-detected images.This fusion enhances the structural details while preserving the most relevant edges for object recognition.
For easier understanding, we will show the entire process of each method in figure \ref{fig3}.

\begin{figure*}[t]
\centering
\includegraphics[width=0.99\linewidth]{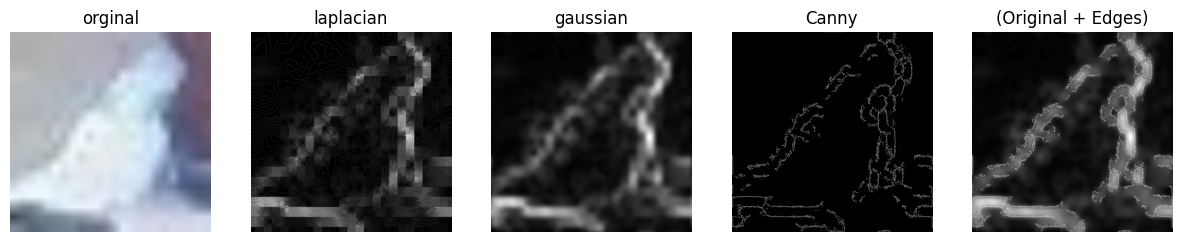}
\caption{ (Follow processing images)  
This is the image transformation process I used the sequential transformations applied to the input image to enhance edge details for object. }
\label{fig3}
\end{figure*} 

\subsection{Model Two-stage}
\subsubsection{Yolov11}
\noindent Real-time object detection has gained significant attention due to its low latency and high applicability. Within the YOLO ecosystem, multiple versions have been developed, with YOLOv12 being the latest. However, due to time constraints, we experimented with YOLOv8, YOLOv9, and even RT-DETRv2—an advanced model known for its superior performance compared to YOLO. After extensive evaluation on our custom dataset, YOLOv11n delivered the highest accuracy, making it the optimal choice for our task.

YOLOv11 retains the overall structure of previous versions, utilizing initial convolutional layers for downsampling input data. However, instead of the traditional C2f block, this model employs the C3k2 block, replacing a large convolution with two smaller ones to enhance computational efficiency while maintaining accuracy . Another significant improvement is the introduction of the C2PSA (Cross-Stage Partial Self-Attention) module, enabling the model to focus on critical regions of an image. This enhancement is particularly beneficial for detecting small or partially occluded objects.

\subsubsection{Efficientnetv2}
\noindent EfficientNetV2 is an improved version of EfficientNet, designed for faster and more efficient image classification. The key enhancements include:
\begin{itemize}
    \item \textbf{Combine MBConv and Fused-MBConv}: As mentioned in the author's paper \cite{effv2} MBConv block often cannot fully make use of modern accelerators. Fused-MBConv layers can better utilize server/mobile accelerators.To overcome this problem, the authors passed both MBConv and Fused-MBConv in the neural architecture search, which automatically decides the best combination of these blocks for the best performance and training speed.
    
    \item \textbf{Progressive Learning}: Taking this hypothesis into consideration, the authors of EfficientNetV2 used Progressive Learning with Adaptive Regularization. The idea is very simple. In the earlier steps, the network was trained on small images and weak regularization. This allows the network to learn the features fast. Then the image sizes are gradually increased, and so are the regularizations. This makes it hard for the network to learn. Overall this method, gives higher accuracy, faster training speed, and less overfitting.

\end{itemize}
\subsubsection{Proposed model}
\noindent Our proposed model consists of three main components: YOLOv11, EfficientNetV2 and Laplacian-based Feature Enhancement.

YOLOv11 is responsible for accurately detecting objects within the frames, particularly for clear objects that are close and unaffected by motion blur. However, for objects that appear farther away or are affected by lighting conditions at night, the detection confidence may be lower. To address this, we incorporate EfficientNetV2 to refine classification accuracy for low-confidence detections made by YOLOv11.

Before passing these low-confidence detections to the classification model, we apply Laplacian-based Feature Enhancement to the cropped object images. This process enhances edge details and sharpness in the input image before classification. Instead of relying solely on YOLOv11's direct output, we extract regions with low confidence, apply a Laplacian filter to enhance edges and important details, and then feed them into EfficientNetV2 for more accurate classification.

By combining these three components, our model effectively leverages the strengths of each method: YOLOv11 ensures fast object detection, EfficientNetV2 optimizes classification accuracy, and Laplacian filtering enhances crucial edge features. This synergy enables the model to generalize better, reduce overfitting, and improve the overall accuracy of the system. To further illustrate our approach, we have designed a pseudo-code representation to explain the logic implemented in our solution.
\begin{algorithm}
\caption{Logic detect fire }\label{alg:yolo}
\begin{algorithmic}
\Require model1, model2, video\_path
\Ensure Processed video with model detect and model classify 

\While{frames in video}
    \For {obj in frames}:
        \If{conf $> 0.6$}
            \State $class\_id \gets \text{int(pred)}$
            \State $boxes.\text{append}([x1, y1, x2 - x1, y2 - y1])$
            \State $scores.\text{append}(\text{float(confidence)})$
            \State $class\_ids.\text{append}(class\_id)$
        \Else
            \State $Crop\_img \gets \text{img}(frame, x1, y1, x2, y2)$
            \State $laplacian\_img \gets \text{Sobel.cal}(Crop\_img)$
            \State $class\_id, conf2 \gets \text{classify}(laplacian\_img)$
            \If{$conf2 < 0.6$}
                \State Discard box
            \Else
                \State $boxes.\text{append}([x1, y1, x2 - x1, y2 - y1])$
                \State $scores.\text{append}(\text{float(conf2)})$
                \State $class\_ids.\text{append}(class\_id)$
            \EndIf
        \EndIf
    \EndFor
    \State Return $(class\_ids, boxes, scores)$
\EndWhile
\end{algorithmic}
\end{algorithm}

\subsection{Patched-YOLO}
To enable the model to detect small and distant objects, we introduce the Patched-YOLO \cite{patched-yolo} method. This approach leverages the spatial characteristics of an image by dividing it into fixed-ratio patches and processing each patch individually.

However, a simple patch-based approach poses a major challenge: objects may be partially cropped, causing the model to miss detections. To address this, we implement an overlapping strategy:
\begin{itemize}
    \item The image is overlapped along both the X and Y axes, ensuring objects appear in multiple patches.
    \item After detection, we apply Non-Maximum Suppression (NMS) to merge overlapping bounding boxes and reconstruct the actual object locations in the original image.
\end{itemize}

This method significantly enhances the model’s ability to detect \textbf{small and distant fire instances}, improving overall detection accuracy.

\section{Experiment}





\subsection{Evaluation Metrics}
\thispagestyle{page5style}

\textbf{Precision} is defined as the ratio of true positive samples (TP) to the total number of samples classified as positive (TP + FP). A high precision indicates that the detected positive samples are highly accurate. When precision equals 1, it means that all detected samples are truly positive, with no false positives.

\begin{equation}
    \text{Precision} = \frac{TP}{TP + FP}
\end{equation}

\textbf{Recall} is defined as the ratio of true positive samples (TP) to the total number of actual positive samples (TP + FN). A high recall indicates a low rate of missing actual positive samples. When recall equals 1, it means that all actual positive samples have been correctly identified.

\begin{equation}
    \text{Recall} = \frac{TP}{TP + FN}
\end{equation}

\textbf{F1-score} is the harmonic mean of precision and recall (assuming both values are nonzero). It ranges within the interval (0,1], where a higher F1-score indicates a better classifier.

\begin{equation}
    F1 = 2 \times \frac{\text{Precision} \times \text{Recall}}{\text{Precision} + \text{Recall}}
\end{equation}

\textbf{Mean Average Precision (MAP)} The Precision-Recall Curve (PRC) describes the behavior and performance of the model.
PRC can be drawn by first sorting the objection detection model output with the confidence scores descending, calculating the cumulative TP, FP, and FN (based on IoU)
at each confidence score threshold. Then, Precision and Recall at each threshold are
computed. Finally, plotting the (precision, Recall) coordinates will result in the PRC.
Average Precision (AP) is a metric to easily summarize the PRC. The AP approximates
the area under the PRC. AP ranges between 0 and 1, where a score of 1 means the model
predicts perfectly all boxes. The larger the metric, the better a model performs across
different confidence score thresholds. The formula for AP is \\
\begin{equation}
    AP = \sum_{k \in n} [r(k) - r(k - 1)] \cdot p(k)
\end{equation}

\begin{align*}
\text{Where:} \\
& k \text{ is the confidence score threshold.} \\
& n \text{  is the set of confidence score thresholds to compute. } \\
& p(k) \text{ is the precision at confidence score threshold } k. \\
& r(k) \text{ is the recall at confidence score threshold } k.
\end{align*}

\textbf{Mean Average Precision (mAP)} is the average AP over all classes. 
\( \text{mAP}@t \) represents mAP computed at the \( t\% \) IoU threshold.

\subsection{Results}
\subsubsection{Detection}
The accuracy of the previous model before we added the dataset was also quite good; however, it was not stable, as shown in Table\ref{acc_old}.
\begin{table}[h]
    \centering
    \renewcommand{\arraystretch}{1.5}
    \begin{tabular}{|c|c|c|c|}
        \hline
         \textbf{} & \textbf{yolov9(RGB-NIR)} &\textbf{yolov7} & \textbf{rt-detr} \\ 
        \hline
         mAP50-95& 0.668-0.598 & 0.51 & 0.65\\
        \hline
    \end{tabular}
    \caption{Accuracy of old model}
    \label{acc_old}
\end{table}
Before retraining the model with our improved dataset, we first tested the old dataset on more modern models.

\begin{figure}[h]
    \centering
    \includegraphics[width=\linewidth]{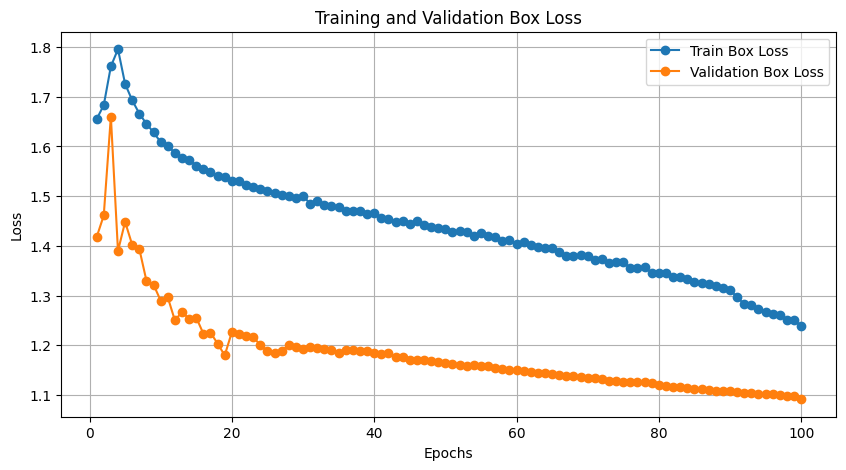}
    \caption{value loss in training process.}
    \label{5}
\end{figure}

\begin{table}[h]
    \centering
    \renewcommand{\arraystretch}{1.5}
    \small 
    \begin{tabular}{|c|c|c|c|}
        \hline
        \textbf{} & \textbf{\shortstack{YOLOv11n \\ (NIR)}} & \textbf{\shortstack{YOLOv11n \\ (RGB)}} & \textbf{\shortstack{RT-DETRv2 \\ (NIR)}} \\ 
        \hline
        mAP50-95 & 0.6 & 0.65 & 0.361 \\
        \hline
    \end{tabular}
    \caption{Accuracy of test model}
    \label{test_model}
\end{table}
As shown in Table \ref{test_model}, the YOLOv11n model has significantly improved accuracy in the fire detection task on old dataset. Therefore, we decided to choose YOLOv11n as the base model for retraining with our improved dataset.
After applying two data augmentation methods, Fusion-GAN and Grayscale conversion, to increase the training dataset, we evaluated the YOLOv11n base model. The resulting accuracy of the model is shown in Table \ref{test1} and Figuer \ref{5}.

\begin{table}[h]
    \centering
    \begin{tabular}{|c|c|c|c|}
        \hline
        Method & mAP@50 & mAP@50:95 & F1-score \\
        \hline
        FIRE-GAN & 0.87 & 0.592 & \\
        Grayscale & \textbf{0.908} & \textbf{0.622} & 0.86 \\
        \hline
    \end{tabular}
    \caption{Comparison of FIRE-GAN and Grayscale methods.}
    \label{test1}
\end{table}





Looking at the results, it can be seen that the proposed model achieves high scores across all three metrics: mAP50, mAP50-95, and F1 score. This indicates that the model misses very few objects, and the detected objects have a high confidence level.

\subsubsection{Classification}

\begin{figure}[h]
    \centering
    \includegraphics[width=\linewidth]{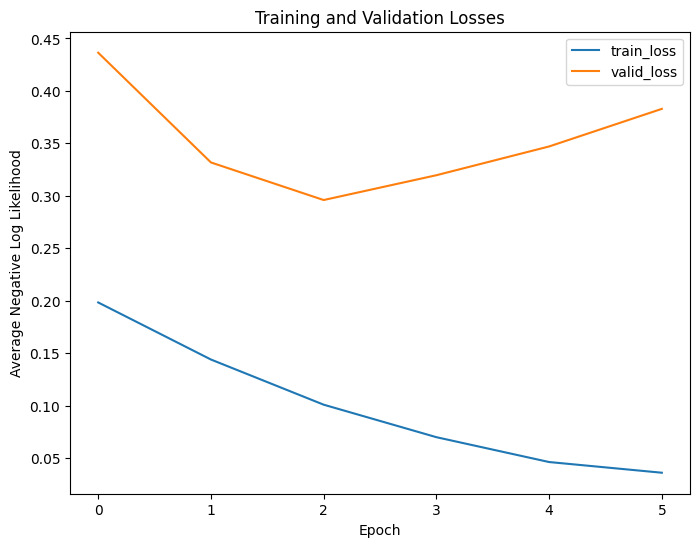}
    \caption{value loss in training process.}
    \label{6}
\end{figure}

For classification, we use log-likelihood for computation. As seen in Figure \ref{6}, the train-loss gradually decreases over epochs. However, the valid-loss increases, indicating that the model is not performing well on the validation set. Therefore, we applied early stopping to select the best model.

Here are the results of our detection and classification models. We are currently working on calculating mAP for the entire process but have encountered some issues. We will do our best to provide the results as soon as possible.


\subsection{Curent limitation}
Based on the model's accuracy, the evaluation results indicate a relatively high performance. However, the model still frequently misclassifies certain objects as fire, especially when the input video or image is blurred or of low quality. This leads to incorrect detections of non-fire objects as fire.

Although the model has improved in terms of accuracy and object detection capabilities, it still struggles with small or distant fire instances. For example, as shown in Figure 5, the model mistakenly identified a non-fire object as fire. Additionally, small fires in the distance, particularly in low-quality video conditions, remain challenging for the model, affecting its overall reliability.

\section{ Conclusion and Future Work}
\noindent In the research, we contribute a pipeline and an accuracy optimization algorithm for fire detection in nighttime infrared camera images. Our approach utilizes two main models, YOLOv11 and EfficientV2, along with various image processing techniques.

Our model has achieved very promising results; however, there are still some limitations when dealing with objects that are very distant and small. Therefore, to make the model more applicable in real-world scenarios, we plan to enhance it by integrating frameworks such as SAHI \cite{SaHi} and patched-yolo to improve the detection of small objects and We will elaborate on these algorithms in the upcoming presentation.

Additionally, we are experimenting with several modifications, including changes in input data, minor network architecture adjustments, and alternative optimization algorithms to accelerate model convergence during training. These improvements aim to enhance the model’s ability to learn edge features more effectively, thereby reducing the need for a second model to reclassify objects when the first model has low confidence.

\bibliographystyle{plain}

\end{document}